\pdfoutput=1

\documentclass[11pt]{article}

\usepackage[preprint]{acl}

\usepackage{times}
\usepackage{latexsym}

\usepackage[T1]{fontenc}

\usepackage[utf8]{inputenc}

\usepackage{microtype}

\usepackage{inconsolata}

\usepackage{graphicx}

\usepackage{hyperref}
\usepackage{url}
\usepackage{xcolor}         
\usepackage{graphicx}
\usepackage{color}
\usepackage{amssymb}
\usepackage{caption}
\usepackage{multirow}
\usepackage{makecell}
\usepackage{longtable} 
\usepackage{array}  
\usepackage{threeparttable}
\usepackage{amsmath}
\usepackage{pifont}
\usepackage[ruled,vlined,linesnumbered]{algorithm2e}
\usepackage{natbib}  
\usepackage{cleveref}
\definecolor{mydarkgreen}{rgb}{0.0, 0.6, 0.4}
\usepackage[utf8]{inputenc}
\usepackage{float}
\usepackage{enumitem}

\title{Theorem-Validated Reverse Chain-of-Thought Problem Generation for Geometric Reasoning}

\author{
 \textbf{Linger Deng\textsuperscript{1}}$^*$,
 \textbf{Linghao Zhu\textsuperscript{1}}$^*$,
 \textbf{Yuliang Liu\textsuperscript{1}}$^\dagger$,
 \textbf{Yu Wang\textsuperscript{2}},
\\
 \textbf{Qunyi Xie\textsuperscript{2}},
 \textbf{Jingjing Wu\textsuperscript{2}},
 \textbf{Gang Zhang\textsuperscript{2}},
 \textbf{Yingying Zhu \textsuperscript{1}},
  \textbf{Xiang Bai\textsuperscript{1}},
\\
\\
 \textsuperscript{1}Huazhong University of Science and Technology,
\\
 \textsuperscript{2}Department of Computer Vision Technology, Baidu Inc,
\\
 \small{
   \textbf{Correspondence:} \href{mailto:email@domain}{lingerdeng, ylliu}@hust.edu.cn
 }
}

\begin{document}
\maketitle
\begin{abstract}
Large Multimodal Models (LMMs) face limitations in geometric reasoning due to insufficient Chain of Thought (CoT) image-text training data. While existing approaches leverage template-based or LLM-assisted methods for geometric CoT data creation, they often face challenges in achieving both diversity and precision. To bridge this gap, we introduce a two-stage Theorem-Validated Reverse Chain-of-Thought Reasoning Synthesis (TR-CoT) framework. The first stage, TR-Engine, synthesizes theorem-grounded geometric diagrams with structured descriptions and properties. The second stage, TR-Reasoner, employs reverse reasoning to iteratively refine question-answer pairs by cross-validating geometric properties and description fragments. 
Our approach expands theorem-type coverage, corrects long-standing misunderstandings, and enhances geometric reasoning. Fine-grained CoT improves theorem understanding and increases logical consistency by 24.5\%. Our best models surpass the baselines in MathVista and GeoQA by 10.1\% and 4.7\%, outperforming advanced closed-source models like GPT-4o. The code is available at \url{https://github.com/dle666/R-CoT}.
\end{abstract}

\let\thefootnote\relax\footnotetext{$^*$Equal contribution.}
\let\thefootnote\relax\footnotetext{$^\dagger$Corresponding author.}

\section{Introduction}
    Large Language Models (LLMs)~\cite{o1systemcard, guo2025deepseek} have revolutionized textual mathematical reasoning through advanced inferential mechanisms. While architectural innovations now enable these models to process multimodal inputs via parameter-efficient vision-language alignment (e.g., GPT-4o~\cite{islam2024gpt}, Gemini~\cite{team2023gemini}), achieving human-competitive VQA performance~\cite{fan2024muffin}, their geometric reasoning remains constrained~\cite{wang2025mv}. This limitation stems from training data dominated by natural scenes, which lack the geometric specificity required for rigorous spatial problem-solving.

    \begin{figure*}[t!]
        \centering
        \includegraphics[width=0.9\linewidth]{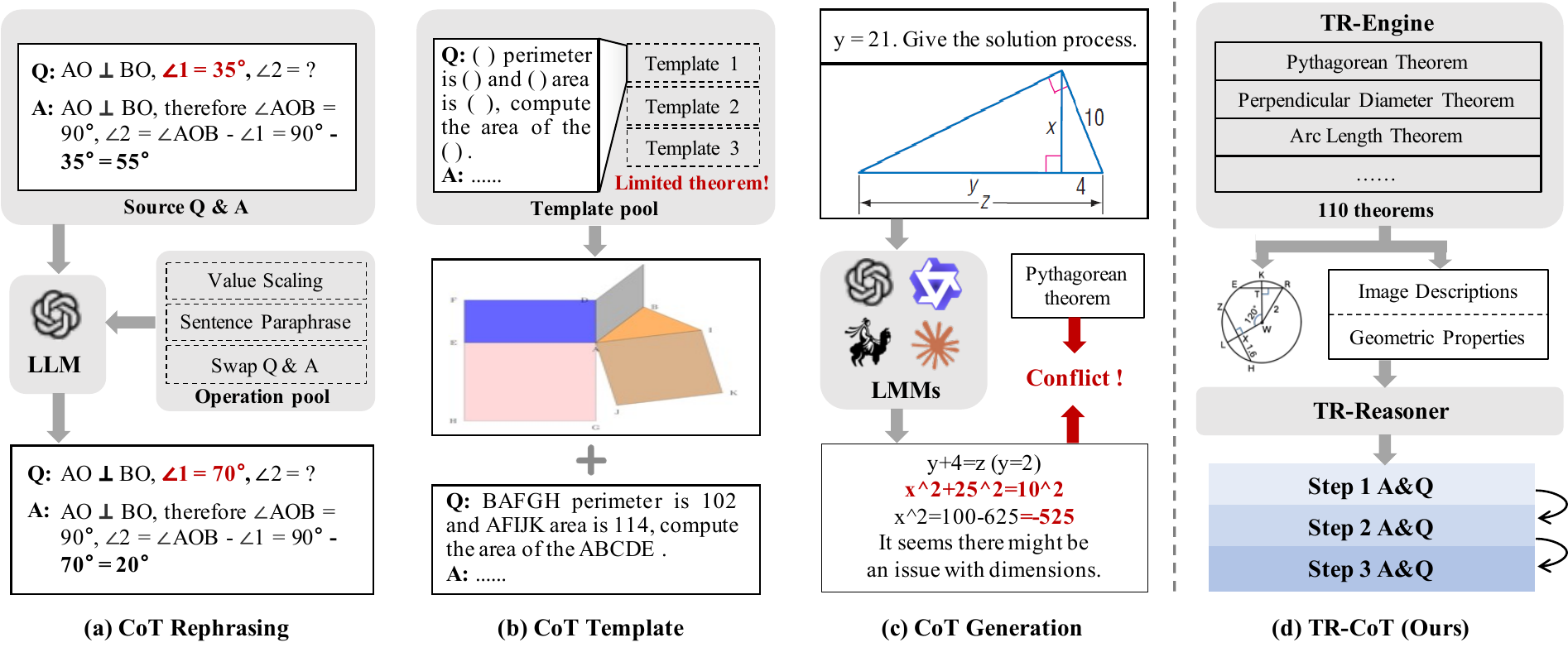}
        \caption{Comparison of TR-CoT with existing CoT data generation approaches. (a) Rephrase existing Q\&A pairs using LLMs, relying on existing CoT data. (b) Generate images and CoT data using pre-defined templates containing a limited number of theorems. (c) Generate CoT using LMMs, where accuracy is limited by the performance of the LMMs. (d) Design the TR-Engine to generate images, corresponding descriptions, and geometric properties from theorems. And input the descriptions and properties into TR-Reasoner to generate reliable CoT Q\&A pairs.}
        \label{fig:motivation}
        \vspace{-8pt}
    \end{figure*}

Current methods for generating geometric reasoning data through Chain-of-Thought (CoT) frameworks face three fundamental limitations.
First, rephrasing approaches \cite{gao2023g} use LLM to transform the CoT format of existing problems, which requires scarce high-quality annotations and domain-specific expertise to ensure theorem consistency (Fig.~\ref{fig:motivation} (a)).
Second, template-based methods \cite{kazemi2023geomverse,zhang2024mavis} generate geometrically oversimplified images by combining predefined polygons in rigid configurations, lacking theorem-aware element interactions, limiting their applicability to advanced reasoning, as shown in Fig.~\ref{fig:motivation} (b).
Thirdly, while LMM-based reasoning ~\cite{peng2024multimath} ensures reasoning diversity, insufficient mathematical priors often lead to incorrect reasoning, e.g., misusing theorems in the wrong situation, leading to logically invalid chains of reasoning(Fig.~\ref{fig:motivation} (c)).

We introduce Theorem-Validated Reverse Chain-of-Thought (TR-CoT), a two-stage framework designed to generate geometric reasoning data and verify logical flows, as shown in Fig.~\ref{fig:motivation} (d). We first develop the theorem-driven image and property generation engine (TR-Engine) to create images paired with geometric properties, ensuring dependencies among elements. Then, TR-Reasoner derives questions from answers by segmenting image descriptions, generating single-step reasoning, and combining them into multi-step reasoning chains. Each step is verified against geometric properties, discarding pairs that violate mathematical rules, ensuring the logical rigor of generated data.

With TR-CoT, we create TR-GeoMM and TR-GeoSup, comprehensive datasets of diverse geometric theorems, which fully leverage CoT information. TR-CoT can bring notable and consistent improvements across a range of LMM baselines such as LLaVA, Qwen, and InternVL. 
Using the recent LMM baselines, we achieve a new performance record in 2B, 7B, and 8B settings for solving geometry problems.
 The main advantages of our method are summarized as follows:
\begin{itemize}[leftmargin=*,noitemsep]
    \item Compared to traditional template-based methods, our approach covers twice the number of theorem types, effectively correcting long-standing theorem misunderstandings in models and enhancing their geometric reasoning.
    \item Generating geometric data with fine-grained CoTs enhances the model’s understanding of theorems, increasing the proportion of logically consistent and clear outputs by 24.5\%.
     \item Our most advanced models achieve a 10.1\% performance gain on MathVista and 4.7\% on GeoQA over the baseline, outperforming advanced closed-source models such as GPT-4o.
\end{itemize}

    \begin{figure*}[t!]
        \centering
        \includegraphics[width=0.9\linewidth]{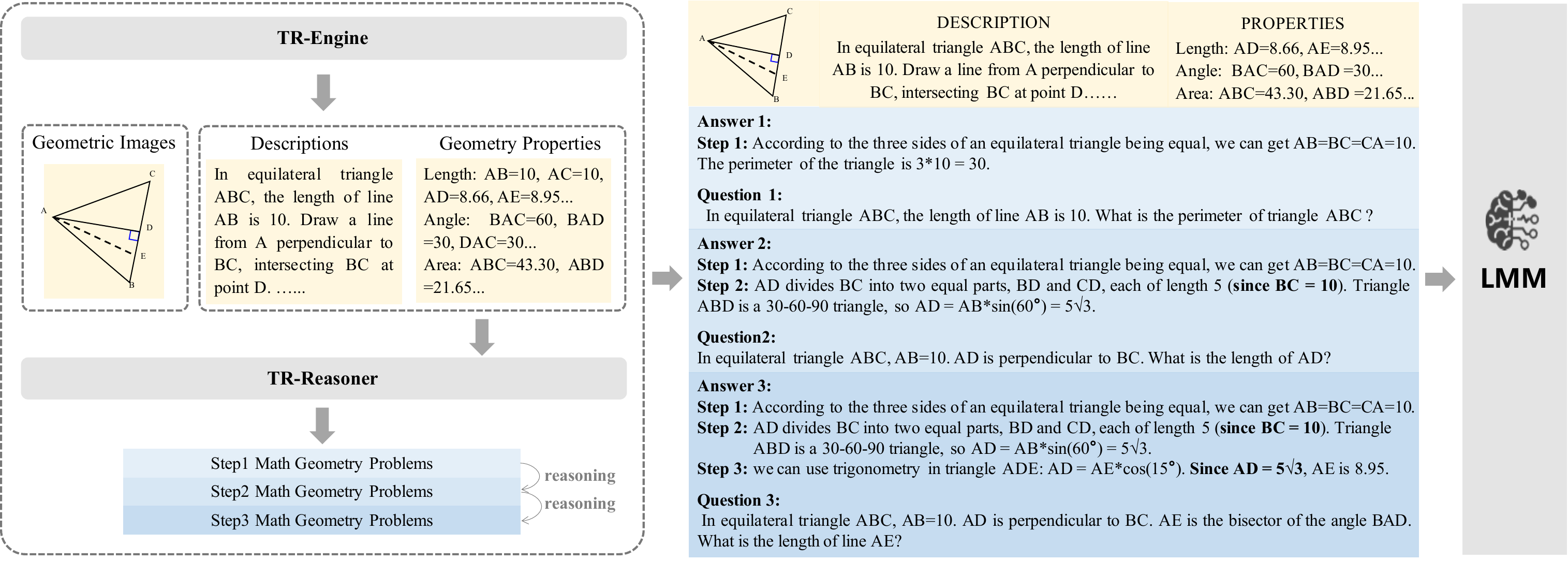}
        \caption{The TR-Engine generates diverse images, corresponding descriptions, and geometric properties step by step based on geometric theorems. Subsequently, the TR-Reasoner is utilized to obtain accurate geometric Q\&A pairs from descriptions and properties.}
        \label{fig:framework}
        \vspace{-8pt}
    \end{figure*}

\vspace{-10pt}
\section{Related Work}
\label{sec:rw}
    \subparagraph{Enhancing Reasoning with CoT in Inference.}
    Chain-of-thought (CoT) prompting has improved reasoning in math tasks. KQG-CoT \cite{liang2023ps} selects logical forms from unlabeled data via CoT-based KBQG. In general math, code-based self-verification \cite{zhou2023} and SSC-CoT \cite{zhao2024s} enhance reliability by combining reasoning with structured knowledge. Other prompting strategies, including PEP \cite{liao2024}, Plan-and-Solve \cite{wang2023plan}, and in-context demonstrations \cite{didolkar2024}, further refine inference. In geometry, visual-symbolic CoT methods \cite{zhao2024a,Hu2024} align reasoning with multimodal representations.

    \subparagraph{Enhancing Reasoning in Geometry Training.}
    Training geometric solvers requires scalable and diverse data. Early symbolic systems (e.g., GeoS \cite{Seo2015}, Inter-GPS \cite{Lu2021}) relied on small benchmarks, while neural approaches like UniGeo \cite{Chen2022} and PGPS9K \cite{zhang2023multi} scaled up with costly manual annotations. Recent methods automate data generation using visual-language models (e.g., G-LLaVA \cite{gao2023}) or code-based engines \cite{kazemi2023,zhang2024mavis}. GeoEval \cite{zhang2024geoeval} provides fine-grained evaluation across diverse reasoning settings. LLM-generated CoT traces \cite{peng2024multimath} offer new avenues for training data synthesis.
    
    Recently, reverse engineering has helped diagnose and refine LLM reasoning. Techniques such as condition-answer swapping \cite{Jiang2024,Weng2023}, error localization \cite{Xue2023}, and prompt optimization \cite{Yuan2024} validate reasoning consistency without model updates. However, they often lack integration into training. Our approach embeds reverse reasoning into CoT generation, producing fine-grained, theorem-aware supervision for model training.

\vspace{-5pt}
\section{Theorem-Validated Reverse Chain-of-Thought}
    There are two key challenges for generating geometry reasoning data: (1) Direct generation of question-answer pairs often leads to errors or unsolvable problems due to oversimplified scenarios. (2) Single-step reasoning processes lack validation of intermediate steps, compromising reliability. 
    
    We propose Theorem-Validated Reverse Chain-of-Thought (TR-CoT), a two-stage framework for creating geometry reasoning data with verified logical flow, as shown in Fig.~\ref{fig:framework}. The pseudo-code of TR-CoT is shown in \Cref{app:Pseudo}.

    1) Stage 1: \textbf{Theorem-Driven Image \& Property Generation.} We collect 110 fundamental geometry theorems (Complete theorems and collection method are shown in \Cref{app:theorems}) and develop \textbf{TR-Engine}, a structured method to generate images paired with textual descriptions and geometric properties (e.g., angles, lengths). 
    Unlike random image generation, TR-Engine guides image generation based on the sampled theorems and enforces dependencies between geometric elements across generation steps. Each current step must operate on the geometric primitives—such as lines, angles, and points—produced in the preceding step.

    2) Stage 2: \textbf{Q\&A Generation with Stepwise Validation.} Using the descriptions and properties from Stage 1, \textbf{TR-Reasoner} generates questions from answers through three steps: First, the image description is divided into logical segments (e.g., “Triangle ABC is isosceles with AB = AC”). An LLM processes these parts step-by-step, generating individual inferences that are then combined into multi-step reasoning chains. Secondly, for each reasoning step, the system creates corresponding questions. For instance, the inference ``$ \angle B =  \angle C$'' generates the question: ``If triangle ABC is isosceles with AB=AC, which angles are equal?'' Finally, all Q\&A pairs are cross-checked against the geometric properties from Stage 1. Pairs violating mathematical rules (e.g., claiming ``$ \angle A = 90^{\circ}$'' for a non-right isosceles triangle) are discarded.

\begin{figure*}[t!]
    \centering
    \includegraphics[width=0.8\linewidth]{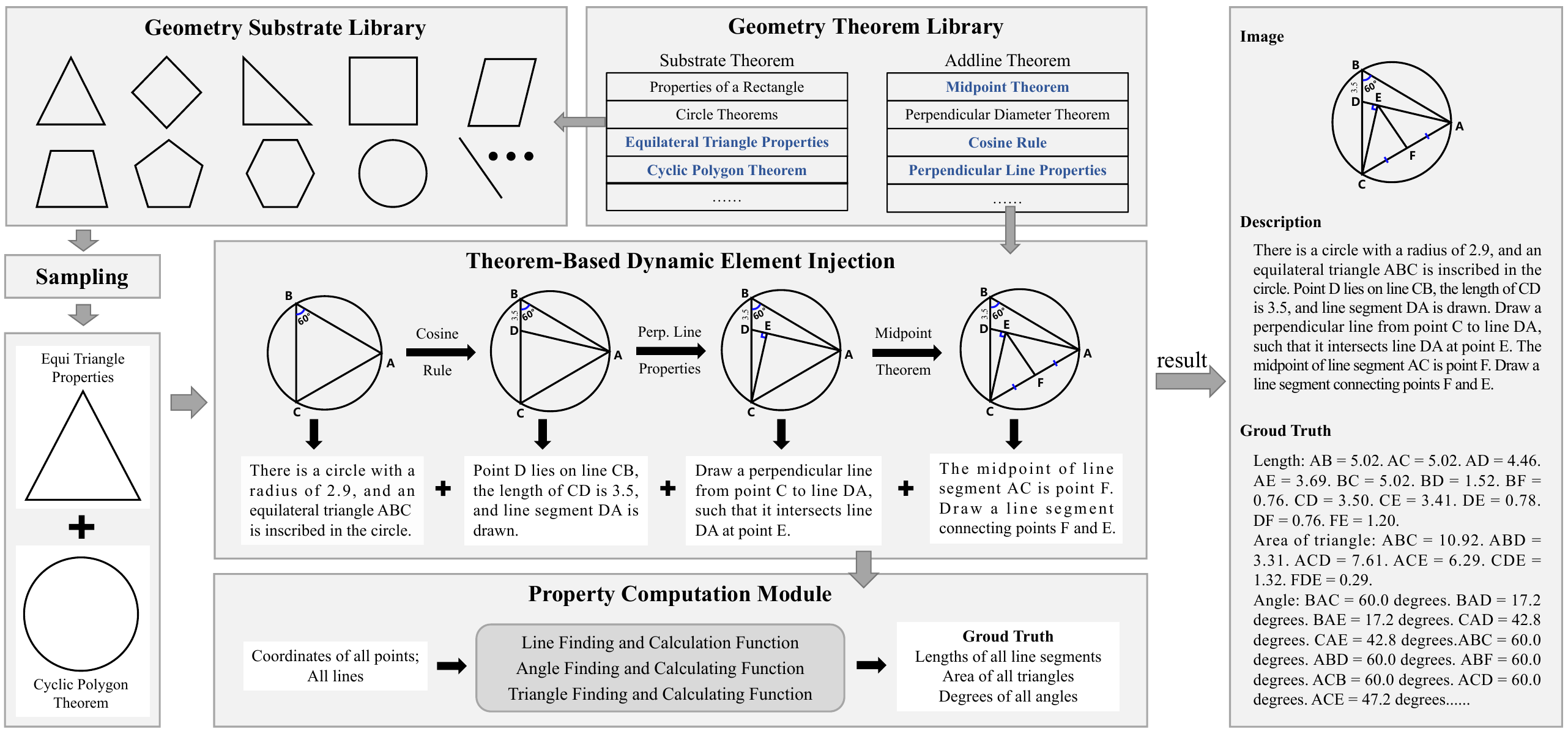}
    \caption{Overview of the TR-Engine. Starting from a Geometric Substrate Library, dynamically injecting elements based on theorems, and integrating a property computation module to enable multi-step geometric reasoning and validation in image generation.}
    \label{fig:engine}
    \vspace{-8pt}
\end{figure*}

\subsection{TR-Engine}
    TR-Engine is a theorem-guided framework for synthesizing geometrically valid images with rich relational structures, corresponding descriptions, and geometric properties. TR-Engine operates through four key components (Fig.~\ref{fig:engine}):

    1) Geometric Theorem Library. The 110 fundamental geometric theorems are classified into substrate-related theorems and line-element-related theorems. During the image generation process, 1 to 3 theorems from each category are sampled to guide the selection of geometric substrates and the addition of line elements.
    
    2) Geometric Substrate Library. We curate 20 fundamental geometric shapes (substrates), such as triangles, circles, and quadrilaterals. Each substrate is paired with a set of relevant geometric theorems and description templates. During image generation, appropriate substrates are selected according to the sampled theorems. The description templates encode geometric conditions (e.g., “In triangle ABC, AB = 5 cm and BC = 6 cm”) to anchor subsequent reasoning steps.

    3) Theorem-Based Dynamic Element Injection. This component strategically injects elements to enable complex reasoning scenarios based on theorem requirements. For example: Adding parallel lines to invoke properties of alternate angles. Introducing auxiliary lines (e.g., medians, altitudes) to create congruent sub-shapes. Such operations expand reasoning opportunities while maintaining geometric validity. In addition, TR-Engine assigns line segment values and angle degrees using exact vertex coordinates, preventing numerical conflicts from geometric constraints.
    
    4) Property Computation Module. As elements are added, the vertex coordinates are used to automatically calculate: Metric properties: Lengths, angles, areas. Relational properties: Parallelism, congruence, symmetry. These properties serve as ground truth for verifying generated Q\&A pairs. Additionally, we perform a visual fidelity check on geometric properties, filtering out distorted images with abnormal vertex spacing (the ratio of the maximum distance to the minimum distance exceeds a threshold) or extreme angles (less than 15 degrees or more than 160 degrees).

    By integrating theorem-driven construction with stepwise validation, TR-Engine ensures images inherently support multi-step geometric reasoning, which is a critical advance over prior generation methods in practice.

    \begin{figure*}[t!]
        \centering
        \includegraphics[width=0.8\linewidth]{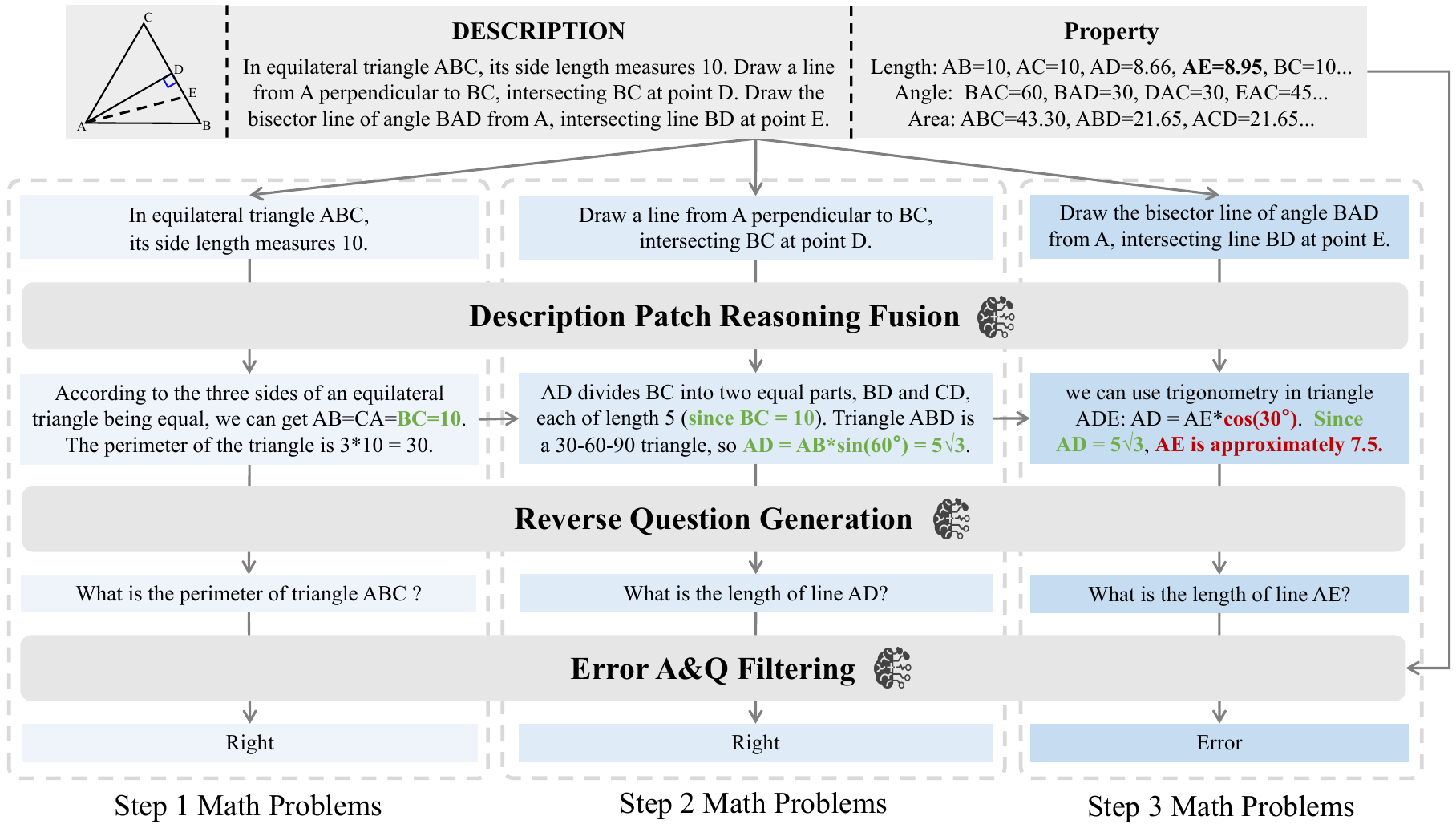}
        \caption{Overview of the TR-Reasoner. Image descriptions are segmented into patches to generate single-step reasoning results. Single-step reasoning results are fused progressively to get multi-step reasoning results. Then questions are generated based on the multi-step reasoning results. Finally,  Q\&A pairs that contradict geometric properties are filtered.}
        \label{fig:cot}
        \vspace{-8pt}
    \end{figure*}

    \subsection{TR-Reasoner}
    Despite advances in LLMs, generating accurate and educationally viable geometric question-answer (Q\&A) pairs remains challenging due to three persistent issues: (1) misapplication of geometric theorems in multi-step proofs, (2) diagram-text misalignment in problem formulation, and (3) inability to maintain answerability constraints during question generation. To address these limitations, we propose the TR-Reasoner to generate theorem-grounded Q\&A pairs through coordinated interaction between geometric properties and structured reasoning chains (Fig.~\ref{fig:cot}).
    
    \paragraph{Description Patch Reasoning Fusion}
    Building on the geometrically valid descriptions from TR-Engine, this module enforces logical coherence through causal dependencies between reasoning steps. Let $D=\{p_1, p_2, ..., p_x\}$ denote the $x$ description patches extracted from an image, where each patch $p_i$ corresponds to a geometrically meaningful component (e.g., ``Circle $O$ with chord AB and tangent CD''). The single-step reasoning $r_i$ for patch $p_i$ is generated through theorem-constrained transformation:
    \vspace{-10pt}
    \begin{equation} 
        r_i = \mathcal{F}_{\text{LLM}}(p_i | r_{ < i}, \mathcal{T}),
    \end{equation}
    where $r_{ < i} = \{r_1, ..., r_{i-1}\}$  represents preceding reasoning states, and $\mathcal{T}$ denotes the applicable theorem set (e.g., intersecting chords theorem for patch $p_i$ describing chord intersections). This chained formulation ensures cumulative reasoning: later steps automatically inherit and extend prior conclusions (e.g., deriving arc lengths after establishing chord congruence).

    \paragraph{Reverse Question Generation} To prevent answerability drift, we implement \textit{answer-constrained reverse generation} rather than open-ended question synthesis. Given a verified reasoning chain $R=\{r_1, r_2, ..., r_n\}$, each step $r_i$ undergoes answerability assessment through a theorem-aware discriminator:

    \vspace{-18pt}
    \begin{equation} 
    f_{aq}(r_i) = \begin{cases} f_q(r_i; \Phi_{\text{geo}}), & \text{if} \mathcal{V}(r_i, G_{\text{props}}) = \text{True} \\ \emptyset, & \text{otherwise} \end{cases} 
    \end{equation}
    
    where $G_{\text{props}}$ denotes geometric properties from TR-Engine (e.g., coordinate-derived lengths), $\mathcal{V}$  performs theorem-based validation (e.g., checking triangle congruence rules), and $f_q$ generates questions using a geometry-specialized LLM with instruction prompt $\Phi_{\text{geo}}$.  
    This approach leverages the granular reasoning steps from the patch reasoning stage to generate theorem-aware Q\&A pairs. 

    \paragraph{Error A\&Q Filtering}
    The final verification stage applies bidirectional cross-validation to ensure Q\&A quality. 
    The forward validation aligns generated answers with deterministic geometric properties computed by TR-Engine's analytical algorithms, removing cases demonstrating: (1) final answer-property discrepancies, and (2) intermediate reasoning inconsistencies with verified properties. 
    The reverse validation identifies ill-posed questions through semantic analysis, excluding those exhibiting answer ambiguity or logical indeterminacy. 
    Both of the validation are conducted through single-round LLM inference, and only Q\&A pairs that satisfy both verifications are reserved. 
    Quantitative analysis revealed four main error patterns that were filtered out: Theorem Violation (36.3\%): incorrect geometric principle application; Metric Discrepancy (24.9\%): numerical inconsistency with problem constraints; Diagram-Text Mismatch (12.2\%): references to non-existent diagram elements; and Answerability Ambiguity (26.6\%): ill-defined problem statements. 
    
    Our proposed filtering mechanism can effectively reduce model hallucination and accumulate errors in previous reasoning steps. Among a sample of 200 generated Q\&A pairs, the framework successfully suppresses reasoning error, reducing overall error rates from 16.0\% (pre-validation) to 5.0\% (post-validation). Showcases of invalid samples in \Cref{app:error_filter_examples}.

\begin{figure*}[t!]
    \centering
    \includegraphics[width=0.95\linewidth]{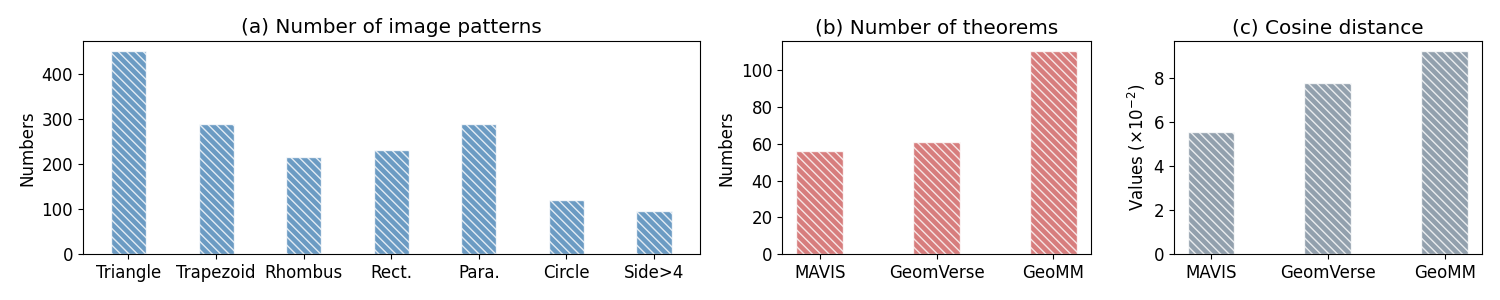}
    \caption{Diversity analysis of TR-GeoMM.}
    \ \label{fig:dataset_three}
    \vspace{-14pt}
\end{figure*}

\vspace{-10pt}
\paragraph{Context-Aware Prompt Engineering}
We deploy an instruction-based context-aware prompting strategy to optimize reasoning. We construct a reasoning instruction template pool containing prototypical geometric problems with a corresponding reasoning process. For each input, 3-4 optimal templates that are most relevant to the theorem and content is selected and integrated into the prompt. Additionally, the pool also contains a series of geometric relationships that are easily misunderstood by LLM. We use the same sample strategy to integrate them into the prompt as well, referred to as Basic Knowledge. The sampled instruction templates and basic knowledge serve as examples to assist the LLM to perform correct reasoning. Such context-aware prompt engineering ensures a relatively ideal reasoning accuracy, improving the efficiency of data generation. More details about the prompt strategy in \Cref{app:aq_prompt}.

\subsection{TR-GeoMM}

\begin{figure}[h]
    \centering
    \includegraphics[width=1\linewidth]{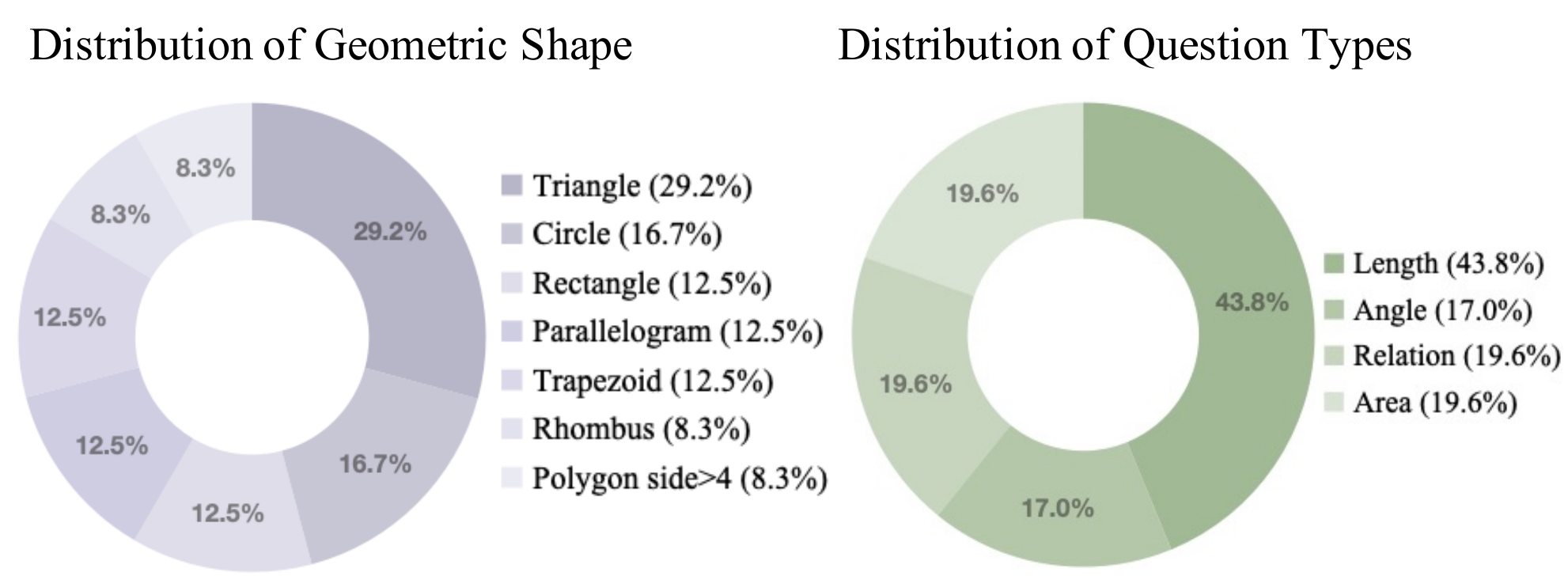}
    \caption{Statistical information about TR-GeoMM.}
    \ \label{fig:dataset}
\end{figure}

Through the TR-CoT pipeline, we construct the TR-GeoMM dataset to enhance LMM's geometric reasoning ability. From 15k figures, we obtain 45k high-quality Q\&A pairs after error filtering, averaging 3.49 questions per figure. Detailed dataset statistics are shown in Fig.~\ref{fig:dataset}.

At the image level, TR-GeoMM covers 20 substrate shapes, mainly triangles, quadrilaterals, and circles. Unlike conventional polygon-based designs, TR-Engine builds figures from lines as primitive elements. It emphasizes key lines with geometric significance, e.g., midlines, angle bisectors, and radii, which frequently appear in theorems. 
As illustrated in Fig.~\ref{fig:dataset_three} (a), 1.7k unique patterns are formed through theorem-guided line combinations, where each addition must interact with existing elements(e.g., a new line's vertex must align with previously generated lines).
At the text level, questions are categorized into four core types: side lengths, angles, areas, and geometric relationships. 
The hierarchical figure construction induces interdependent questions, where earlier solutions serve as prerequisites for subsequent ones. This subproblem design supports step-by-step learning of geometric concepts and reasoning. 
As shown in Fig.~\ref{fig:dataset_three} (b), TR-GeoMM contains a theorem repository twice as large as existing synthetic datasets (MAVIS and GeomVerse). 
Furthermore, Fig.~\ref{fig:dataset_three} (c) demonstrates superior data diversity through higher Q\&A pair cosine distances. More information is provided in \Cref{app:geo_dataset} and \Cref{app:compare}.

\subsection{TR-GeoSup}

TR-CoT can not only generate reliable CoT geometric data but also be used to augment existing datasets. Real-world geometry CoTs often include key intermediate steps rich in problem-solving insights, yet these are typically implicit or oversimplified, relying on human prior knowledge. This lack of explicit reasoning may hinder model learning due to limited background knowledge and inference capability. Leveraging the TR-CoT pipeline, we decompose the original CoT process into explicit theorem-aware steps, then reverse generate new Q\&A pairs with TR-Reasoner.

Specifically, our augmentation involves three steps: generating a comprehensive multi-step analysis of the geometric figure, segmenting it into essential problem-solving steps, and creating new Q\&A pairs for each step. These fine-grained Q\&A pairs explicitly guide the model with theorems and knowledge implicitly expressed in the original data, leading to improvement in comprehension and reasoning abilities. We applied TR-Reasoner to the GeoQA dataset, producing the TR-GeoSup dataset with 20k new entries. The final TR-GeoSup dataset does not contain the original GeoQA data. Examples of TR-GeoSup are shown in \Cref{app:geosup}.

During the augmentation, LLM receives the original question and its corresponding CoT answer to produce a more complete analysis, supplementing missing theorems and steps not explicitly stated in the original CoT. We sampled 200 examples from both the analysis and Q\&A generation stages and observed no errors, confirming the reliability of our design. To streamline the data generation process, we did not introduce additional independent validation. After generation, 10\% of the data was manually reviewed and corrected.



\vspace{-10pt}
\section{Experiments}
\subsection{Setup}
We train multiple LMMs~\citep{wang2024qwen2, liu2024visual, chen2024internvl} using existing geometric instruction datasets ~\citep{chen2021geoqa, gao2023g} and our TR-CoT generated data (TR-GeoMM and TR-GeoSup). Both the projected linear layer and the language model are trainable. The models are trained for two epochs with a batch size of 128 on $16\times$ 64G NPU, and learning rate set to 5e-6. For evaluation, we assess these models on the geometry problem solving on the testmini set of MathVista ~\citep{lu23mathvista} and GeoQA ~\citep{chen2021geoqa} following~\citet{gao2023g}. Top-1 accuracy serves as the metric, with predictions and ground truth evaluated via ERNIE Bot 4.0. Ablation experiments were done on Intern-VL-2.0-8B.

\subsection{Ablation Study}
\paragraph{Data generating procedures.}
To evaluate the contributions of TR-CoT components, we construct ablated variants by removing specific modules, as summarized in Tab.~\ref{rcot}. Each variant is used to generate training data, and the resulting models are evaluated on MathVista and GeoQA.
Generating Q\&A pairs from descriptions yields better performance than from images, with gains of 5.3\% on MathVista and 6.3\% on GeoQA. Incorporating reverse generation further improves accuracy by 2.9\% and 2.6\% on the two datasets, respectively.
The full TR-CoT pipeline achieves the best performance, confirming the effectiveness of each component.

\begin{table}[h]
\small
\centering
\caption{Ablation study on the data generating procedures. `Description' represents generation based on descriptions. `Reverse' represents generating reasoning followed by reverse question generation. `Filter' represents filtering errors based on geometric properties.}
\renewcommand\arraystretch{1.1}
\setlength{\tabcolsep}{2.1mm}{
\begin{tabular}{ccc|ll}
\hline
\multicolumn{3}{c|}{Configurations}                    & \multicolumn{1}{c}{}                            & \multicolumn{1}{c}{}                        \\ \cline{1-3}
Description    & Reverse      & Filter & \multicolumn{1}{c}{\multirow{-2}{*}{MathVista}} & \multicolumn{1}{c}{\multirow{-2}{*}{GeoQA}} \\ \hline
\ding{55}       & \ding{55}             & \ding{55}          & \quad 55.3                & \quad 44.2             \\
\checkmark      & \ding{55}             & \ding{55}          & \quad 60.6                    & \quad 50.5                 \\
\checkmark      & \checkmark           & \ding{55}       & \quad 63.5                      & \quad 53.1                \\
\checkmark      & \checkmark          & \checkmark       & \quad 64.4                       & \quad 54.0                  \\ \hline
\end{tabular}
}
\label{rcot}
\end{table}

\begin{figure*}[t!]
    \centering
    \includegraphics[width=1\linewidth]{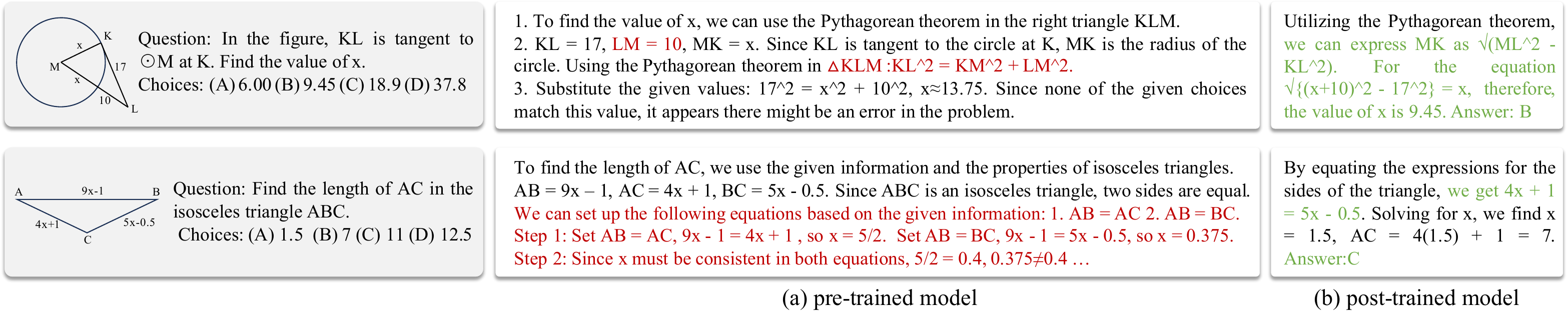}
    \caption{Comparison of model problem solving before and after training.}
    \label{fig:math_example}
    \vspace{-8pt}
\end{figure*}

\vspace{-10pt}
\paragraph{Separate validity of synthetic and augmented data.}
We evaluated the impact of the TR-GeoSup and TR-GeoMM datasets on model performance, as shown in Tab.~\ref{rcot_dataset}. Training with TR-GeoSup improved performance by 1.4\% on MathVista and 7.9\% on GeoQA compared to the baseline. Combining GeoQA with TR-GeoSup improves performance by 2.9\% on MathVista and 3.9\% on GeoQA compared to GeoQA alone, indicating their complementarity.
It suggests TR-GeoSup effectively enhances in-domain performance with better extracted knowledge. A deeper understanding of knowledge may contribute to improved generalization on mixed out-of-domain datasets. 

\begin{table}[h]
\small
\centering
\caption{Ablation study on the TR-CoT generated data.}
\renewcommand\arraystretch{1.1}
\setlength{\tabcolsep}{1.3mm}{
\begin{tabular}{ccc|ll}
\hline
\multicolumn{3}{c|}{Configurations}                    & \multicolumn{1}{c}{}                            & \multicolumn{1}{c}{}                        \\ \cline{1-3}
GeoQA    & TR-GeoSup      & TR-GeoMM & \multicolumn{1}{c}{\multirow{-2}{*}{MathVista}} & \multicolumn{1}{c}{\multirow{-2}{*}{GeoQA}} \\ \hline
\ding{55}       & \ding{55}           & \ding{55}           & \quad 63.0                      & \quad 52.4                 \\
\checkmark      & \ding{55}           & \ding{55}           & \quad 64.9                    & \quad 64.8               \\
\ding{55}       & \checkmark          & \ding{55}           & \quad 64.4                      & \quad 60.3                 \\
\ding{55}       & \ding{55}           & \checkmark          & \quad 64.4                      & \quad 54.0                 \\
\checkmark      & \checkmark          & \ding{55}           & \quad 67.8                      & \quad 68.7                 \\
\checkmark      & \ding{55}           & \checkmark          & \quad 65.4                      & \quad 67.9                 \\
\checkmark      & \checkmark          & \checkmark          & \quad 68.3                      & \quad 69.0                   \\ \hline
\end{tabular}
}
\label{rcot_dataset}
\end{table}
\vspace{-10pt}

Second, training with TR-GeoMM improved performance by 1.4\% on MathVista and 1.6\% on GeoQA, confirming the strong generalization of TR-CoT synthetic data to real data. Moreover, joint training with GeoQA further improved performance, highlighting the effectiveness of synthetic data in supplementing real data.
Finally, when jointly training on all three datasets 
(GeoQA, TR-GeoSup and TR-GeoMM). The model achieved the best performance, with improvements of 5.3\% on MathVista and 6.6\% on GeoQA over the baseline. These results support that TR-CoT-generated data compensate for the limitations of existing datasets and enhance the model’s reasoning capability.

\paragraph{Compared with other synthesis datasets.}
We train InternVL-2.0-8B using TR-GeoMM and two recent synthetic datasets for geometric problems, \emph{i.e.} MAVIS (synthesis part)~\citep{zhang2024mavis} and GeomVerse~\citep{kazemi2023geomverse}, as summarized in Tab.~\ref{compare_syn}. Compared to the baseline, models trained with GeomVerse or MAVIS show a slight performance gain on GeoQA and a decline on MathVista, both lower than TR-GeoMM. We attribute this to the limited diversity of image and Q\&A pairs in these datasets, which benefits the simpler distribution of GeoQA but struggles with the diverse distributions in MathVista. In contrast, TR-GeoMM, with its diverse image and Q\&A pairs, improves performance on both datasets.

\begin{table}[h]
\small
\caption{Compared with other synthesis datasets.}
\centering
\renewcommand\arraystretch{1.1}
\setlength{\tabcolsep}{4mm}{
\begin{tabular}{c|ccc}
\hline
Dataset  & MathVista & GeoQA\\  \hline
   /        & 63.0  & 52.4 \\
GeomVerse(9k)    & 58.2  & 53.6 \\
MAVIS(sample 48k)   & 57.2  & 53.2 \\ 
TR-GeoMM(45k)  & 64.4  & 54.0  \\ 
TR-GeoMM(sample 9k)  & 63.0  & 55.6  \\ \hline
\end{tabular}
}
\label{compare_syn}
\end{table}
\vspace{-10pt}

\subsection{Comparison with Previous State-of-the-Art}
With the proposed method, we train three specialized models for geometry problem solving named TR-CoT-InternVL-2.0-2B, TR-CoT-Qwen2.5-VL-7B, and TR-CoT-InternVL-2.5-8B on the joint dataset of Geo170K and TR-CoT-generated data (TR-GeoMM and TR-GeoSup). We compare our models with both general and mathematical LMMs on the geometry problems from testmini set of MathVista and the test set of GeoQA. 
As shown in Tab.~\ref{result}, TR-CoT-InternVL-2.5-8B outperforms GPT-4o by 17.3\% on MathVista and TR-CoT-Qwen2.5-VL-7B outperforms GPT-4o by 17.8\% on GeoQA. Compared to mathematical LMMs, TR-CoT-InternVL-2.5-8B maintains a 11.1\% lead on MathVista, and TR-CoT-Qwen2.5-VL-7B achieves a 12.5\% advantage on GeoQA. 
For performance analysis on more baselines, please refer to \Cref{app:Effectiveness} and Tab.~\ref{Effectiveness_result}.

\begin{table}[h]
\small
\centering
\caption{Top-1 Accuracy (\%) on geometry problem solving on the testmini set of MathVista and the GeoQA test set. * represents the results from the existing papers.}
\renewcommand\arraystretch{1.1}
\setlength{\tabcolsep}{0.5mm}{
\begin{tabular}{c|c|c}
\hline
Model                       & MathVista         & GeoQA                          \\ \hline
\multicolumn{3}{c}{Closed-source LMMs} \\ \hline
GPT-4o~\citep{islam2024gpt}                              & 60.6         & 61.4              \\
GPT-4V                                                  & \ 51.0*      & \ 43.4*                 \\
Gemini Ultra  ~\citep{team2023gemini}                   & \ 56.3*      & -                 \\ \hline
\multicolumn{3}{c}{Open-source LMMs} \\ \hline
LLaVA2-13B ~\citep{liu2024visual}                 & \ 29.3*      & \ 20.3*     \\
mPLUG-Owl2-7B~\citep{ye2024mplug}                        & 25.5         & 21.4      \\
Qwen-VL-Chat-7B~\citep{bai2023qwen}                      & 35.6         & 26.1     \\ 
Monkey-Chat-7B~\citep{li2024monkey}                      & 24.5         & 28.5    \\
Deepseek-VL-7B~\citep{lu2024deepseek}                    & 34.6         & 33.7     \\ 
InternVL-2.0-2B ~\citep{chen2024internvl}         & 46.2         & 38.2    \\
InternLM-XC2-7B~\citep{zhang2023internlm}         & 51.4         & 44.7        \\
InternVL-1.5-20B~\citep{chen2024far}                & 60.1         & 49.7        \\ 
Qwen2-VL-7B ~\citep{wang2024qwen2}        & 55.1         & 55.7    \\
InternVL-2.0-8B ~\citep{chen2024internvl}        & 65.9         & 56.5    \\ 
InternVL-2.5-8B   ~\cite{chen2024expanding}         & 67.8         & 59.0     \\
Qwen2.5-VL-7B ~\citep{wang2024qwen2}           & 71.6         & 74.5     \\
\hline
\multicolumn{3}{c}{Open-source Mathematical LMMs} \\ \hline
UniMath ~\cite{liang2023unimath}  & -      & \ 50.0*   \\
Math-LLaVA-13B ~\citep{shi2024math}                      & \ 56.5*      & 47.8         \\
G-LLaVA-7B~\citep{gao2023g}                              & \ 53.4*      & \ 62.8*       \\
MAVIS-7B~\citep{zhang2024mavis}                          & -            & \ 66.7*        \\ 
PUMA-Qwen2-7B ~\cite{zhuang2024math}  & \ 48.1*  & -   \\
MultiMath-7B ~\cite{peng2024multimath}  & \ 66.8*   & -  \\\hline
TR-CoT-InternVL-2.0-2B     & 56.3           &  63.4 \\
TR-CoT-Qwen2.5-VL-7B      &  74.5          &  \textbf{79.2} \\
TR-CoT-InternVL-2.5-8B      &  \textbf{77.9}          &  76.7 \\
\hline
\end{tabular}
  }
\label{result}
\end{table}

\vspace{-10pt}
\section{Discussion}

Fig.~\ref{fig:math_example} highlights consistent improvements: post-trained models produce concise, logical CoTs with accurate conclusions, demonstrating robust geometric understanding. Pre-trained models show recurring errors (e.g., misdefining isosceles triangles as having two equal sides), reflecting foundational gaps in theorem comprehension. Our approach trains models on diverse theorems with structured reasoning, addressing these errors and enhancing general geometric problem-solving.

\begin{figure}[t!]
    \centering
    \includegraphics[width=1\linewidth]{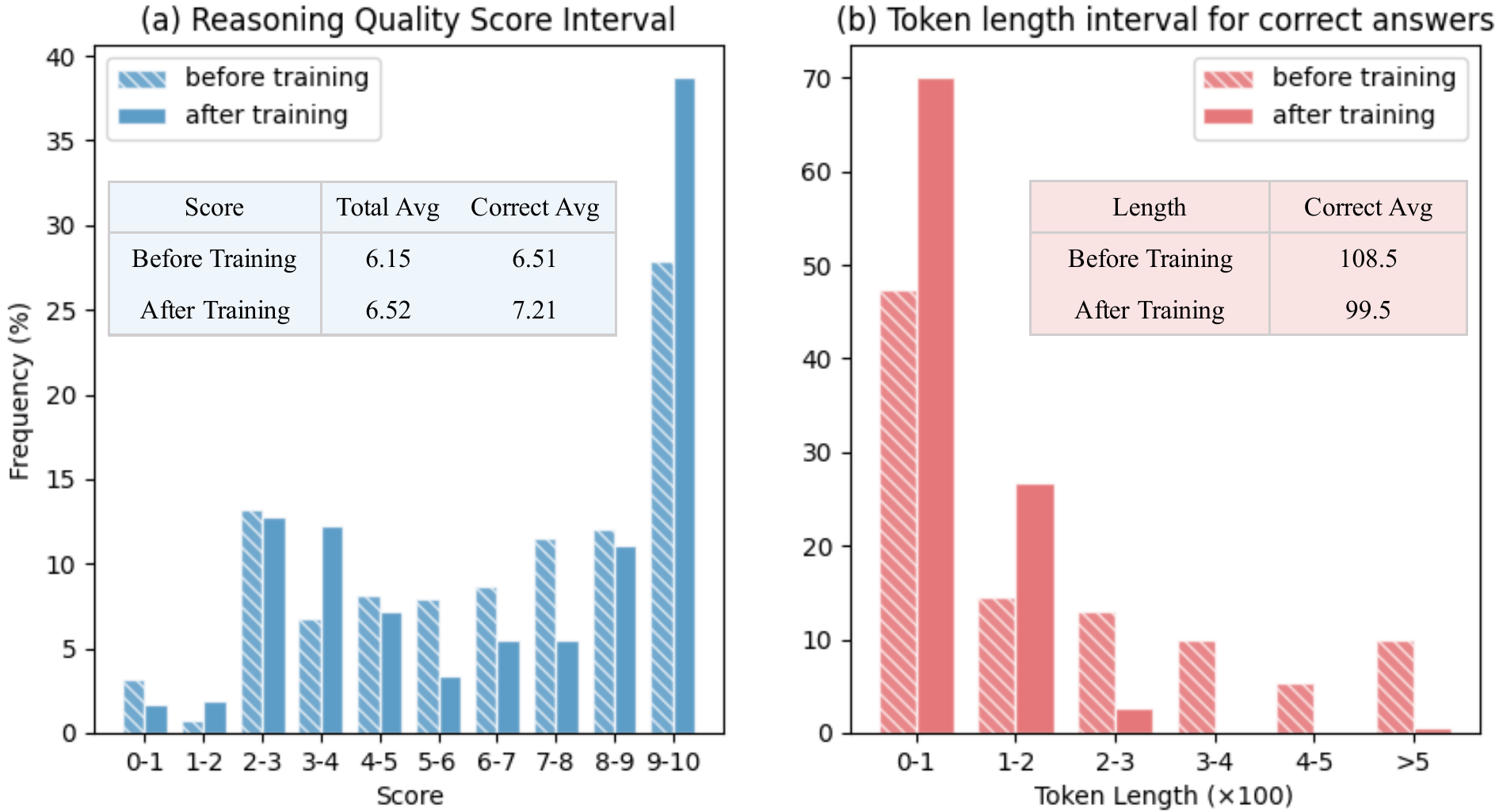}
    \caption{Comparison of model output quality and token length before and after training.}
    \label{fig:rcot_advance}
\end{figure}

We use DeepSeek R1 and ERNIE Bot 4.0 to quantitatively evaluate model outputs before and after training, focusing on logical consistency, clarity, and lack of ambiguity (see \Cref{app:cot_quality} for detailed information). We use the average score of the two models as the final score. As shown in Fig.~\ref{fig:rcot_advance} (a), the total mean score increased by 0.37 after training, the mean score for correct answers increased by 0.70, and outputs with scores of 8 or higher increased by 24.5\%. We attribute these improvements to TR-CoT's explicit focus on the reasoning process, where step decomposition enhances the model's logical consistency and rigor.

We further compare the token usage for correct answers before and after training. As shown in Fig.~\ref{fig:rcot_advance} (b), the model after training requires fewer tokens on average, with the percentage of correct answers within 200 tokens increasing by 35\%. We assume this improvement results from the data diversity, which enables the model to find more efficient solutions across different theorems, while a deeper understanding of the theorems allows for more concise reasoning.

\section{Conclusion}
We propose TR-CoT, a novel theorem-based reverse generation pipeline that enhances theorem coverage and supports fine-grained theorem understanding in geometric datasets. Models trained on TR-CoT data demonstrate a significant improvement in geometric problem solving with more concise and rigorous reasoning. We will extend this approach to other mathematical domains to further analyze the impact of theorem mastery on problem-solving, offering insights for future research.

\section*{Limitations}

For our method, one major constraint is that there is still room for further improvement in the generation efficiency. The overall efficiency can be divided into time efficiency and data efficiency. First, in our process, LLM is called multiple times for reasoning generation. The limited reasoning speed of LLM becomes the bottleneck of time efficiency. In addition, although we have adopted various methods to improve the reasoning accuracy of LLM, due to the limitations of model performance, there is still a certain proportion of errors in the direct output of the model. We observe that about 10\% of the direct output is deleted in the Error A\&Q Filtering stage.



\bibliography{custom}
\clearpage
\appendix

\section{Pseudo Code}
\label{app:Pseudo}
We have written pseudo-code for the overall flow of TR-CoT, the details of which are given in Algor.~\ref{r-cot algorithm}. 

    \begin{algorithm}
    \SetAlgoLined
    \DontPrintSemicolon
    \SetNoFillComment
    \footnotesize
    \KwIn{Geometry substrates sampling rounds $n$, plot function $f$, image-description pair sets $\mathcal{S}$, line sampling rounds $k$, geometric property calculation module $\mathcal{V}$, large language model $\mathcal{M}$}
    \KwOut{Generated Image $\mathcal{I}$, Description $\mathcal{D}$, Geometric Properties $\mathcal{T}$, Question $\mathcal{Q}$; Answer $\mathcal{A}$}
    
    Initialization: $\mathcal{I} \leftarrow \emptyset$, $\mathcal{D} \leftarrow \emptyset$, $\mathcal{T} \leftarrow \emptyset$, vertex coordinate $\mathcal{C} \leftarrow \emptyset$, $r_{s} \leftarrow \emptyset$\; 
    \For{i $\leftarrow \text{1}$ to $n$}{
        Sample geometry substrate $\mathcal{G}_{i}$ and description $\mathcal{D}_{i}$ from image-description pair sets $\mathcal{S}$ \;
        Refresh $\mathcal{I}$ using plot function:
        $\mathcal{I} \leftarrow f(\mathcal{I}, \mathcal{G}_i)$ \;
        Refresh corresponding description: $\mathcal{D} \leftarrow \mathcal{D} \cup \mathcal{D}_i$ \;
        Refresh vertex coordinate: $\mathcal{C} \leftarrow \mathcal{C} \cup \mathcal{C}_i$ \;
    }
    \For{j $\leftarrow \text{1}$ to $k$}{
        Select line drawing position $\mathcal{P}_j$ \;
        Draw line and label length: $\mathcal{I} \leftarrow f(\mathcal{I}, \mathcal{P}_j)$ \;
        Refresh corresponding description: $\mathcal{D} \leftarrow \mathcal{D} \cup \mathcal{P}_j$ \;
        Refresh vertex coordinate: $\mathcal{C} \leftarrow \mathcal{C} \cup \mathcal{C}_i$ \;
        \If{j = k}{
            Calculate all angle information $\mathcal{R}$ \;
            Draw angles and label degrees: $\mathcal{I} \leftarrow f(\mathcal{I}, \mathcal{R})$ \;
            Refresh corresponding description: $\mathcal{D} \leftarrow \mathcal{D} \cup \mathcal{R}$ \;       
        }
    }
    Refresh Geometric Properties: $\mathcal{T} \leftarrow \mathcal{V}(C)$ \;
    
    Produce single-step reasoning result $r_{c}$ using prompt $P_{s}$: $r_{c} \leftarrow \mathcal{M}(\mathcal{D}, P_{s})$ \;
    Generate answer $A_{e}$ and its corresponding question $Q_{e}$ using prompt $P_{q}$: $A_{e}, Q_{e} \leftarrow \mathcal{M}(r_{c}, P_{q})$ \; 
    Filtering for correct answer $A$ and its corresponding question $Q$  using prompt $P_{e}$: $A, Q \leftarrow \mathcal{M}(A_{e}, Q_{e}, T, P_{e})$ \; 
    Return: $\mathcal{I}$, $\mathcal{D}$, $\mathcal{Q}$, $\mathcal{A}$
    \caption{Pseudo-code of TR-CoT}
    \label{r-cot algorithm}
    \end{algorithm}
    
\section{Details of prompt in TR-Reasoner}
\label{app:aq_prompt}
\begin{figure}[t!]
    \centering
    \includegraphics[width=0.9\linewidth]{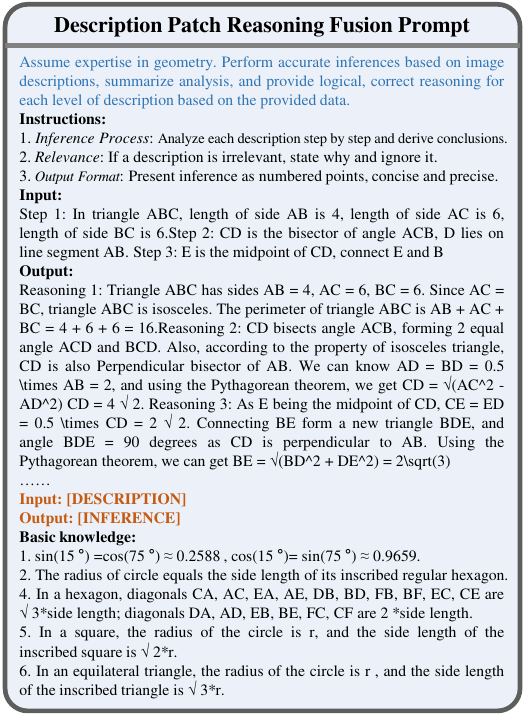}
    \caption{The prompt of the Description Patch Reasoning Fusion.}
    \label{fig:prompt1}
\end{figure}

\begin{figure}[t!]
    \centering
    \includegraphics[width=0.9\linewidth]{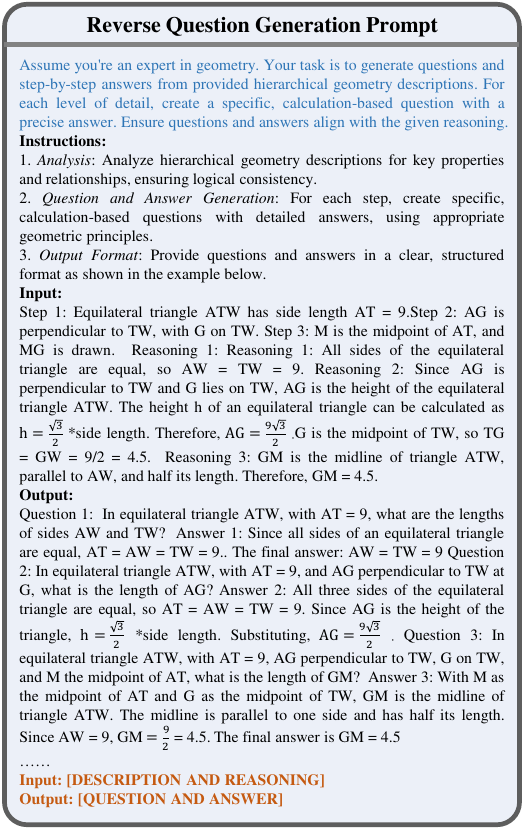}
    \caption{The prompt of the Reverse Question Generation.}
    \label{fig:prompt2}
\end{figure}

\begin{figure}[t!]
    \centering
    \includegraphics[width=0.9\linewidth]{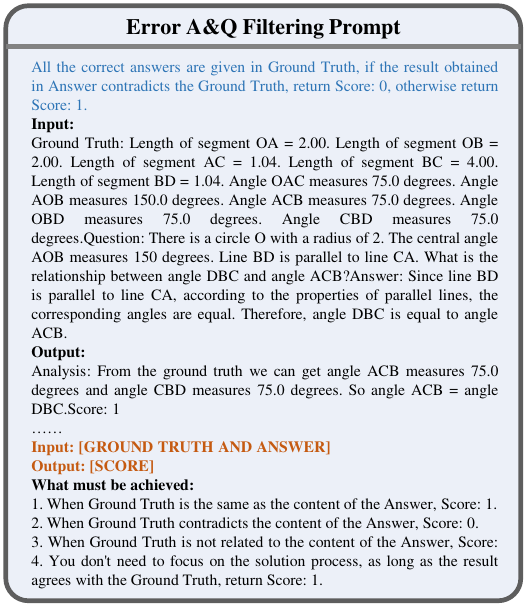}
    \caption{The prompt of the Error A\&Q Filtering.}
    \label{fig:prompt3}
\end{figure}

We used ERNIE Bot 4.0 to implement TR-Reasoner. We describe the prompts used in TR-Reasoner, including the prompts for the Description Patch Reasoning Fusion (Fig.~\ref{fig:prompt1}), the Reverse Question Generation (Fig.~\ref{fig:prompt2}), and the Error A\&Q Filtering (Fig.~\ref{fig:prompt3}). In these figures, the texts in blue represent the Task Description, while the texts in orange represent the input information. Each prompt includes three contextual examples, and we show only one of them, with the remaining examples replaced by ellipses. In addition to the examples, some prompts also include an instruction section that specifies more detailed requirements, some incorporate additional basic knowledge, and others outline more specific goals that must be achieved.

In preliminary experiments, we observed that LLMs often failed to accurately interpret certain geometric relationships. To systematically identify such issues, we selected 50 representative instances per geometric substrate from the TR-GeoMM dataset and applied the TR-Reasoner framework for Patch Reasoning. We analyzed the most frequently misinterpreted relationships and formalized their correct representations into a base knowledge library. During formal generation, prompts are dynamically constructed by retrieving relevant geometric relationships from this library based on the target substrate.

\section{More information of TR-GeoMM}
\label{app:geo_dataset}
\begin{figure}[t!]
    \centering
    \includegraphics[width=0.9\linewidth]{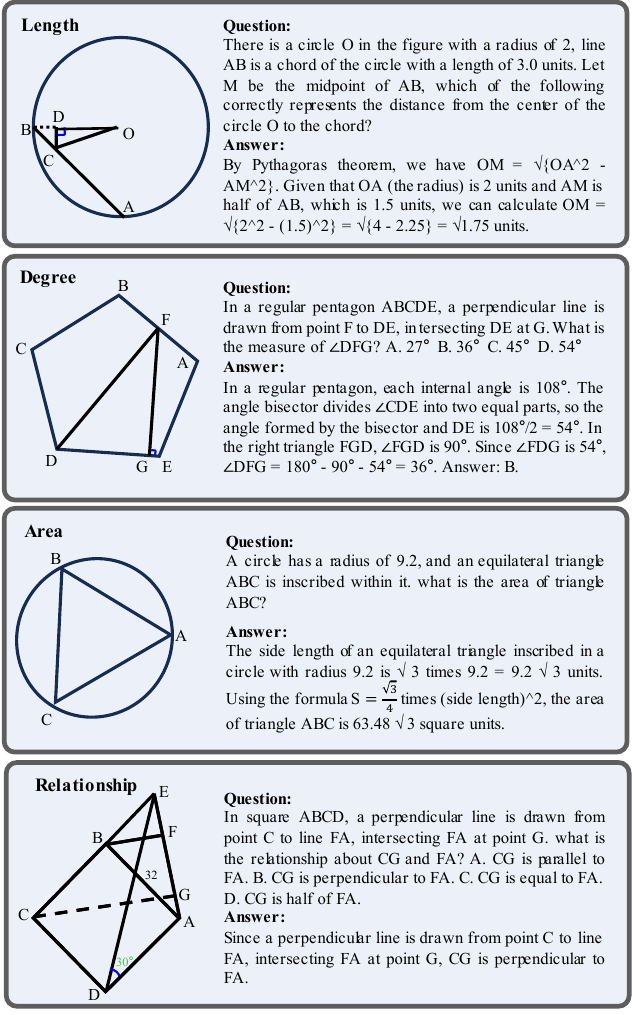}
    \caption{Examples of TR-GeoMM dataset.}
    \label{fig:qa_example}
\end{figure}

Through the TR-CoT, we construct a high-quality geometric dataset, TR-GeoMM. In Fig.~\ref{fig:qa_example}, we provide a detailed overview of specific cases from TR-GeoMM. These cases demonstrate the variety of mathematical geometry question types covered by TR-GeoMM, including solving for lengths, angles, areas, and geometry elemental relations. Each of these categories is critical for improving the geometric reasoning ability of LMMs.

For Cosine distance based data diversity, we first randomly sample 5000 instances from each dataset(MAVIS, GeomVerse, and TR-GeoMM), then we encode the instances into embedding features using pretrained BERT model~\cite{devlin2018bert}. Finally, we calculate the average cosine distance of each dataset using the BERT output features. Higher distance score indicates better diversity, and our TR-GeoMM has the highest distance score among the three datasets.

We further report the computation and generation cost of the generation pipeline here. Geometric images, descriptions, and properties are all generated simultaneously by the TR-Engine, our Python-built graphics rendering engine. A total of 15,000 geometric images were generated during the geometric image generation phase, and it could be done in approximately 10 minutes by a 12th-gen Intel Core i7-12700. TR-Reasoner, our LLM-based module for Patch Reasoning, Q\&A generation, and error filtering, runs via cloud APIs with 32 parallel processes. These stages take approximately 6 hours in total (~2 hours per stage). As a one-time cost, the generated data can be reused across model training tasks.

\section{Qualitative examples of filtered errors}
\label{app:error_filter_examples}
\begin{figure*}[t!]
    \centering
    \includegraphics[width=0.9\linewidth]{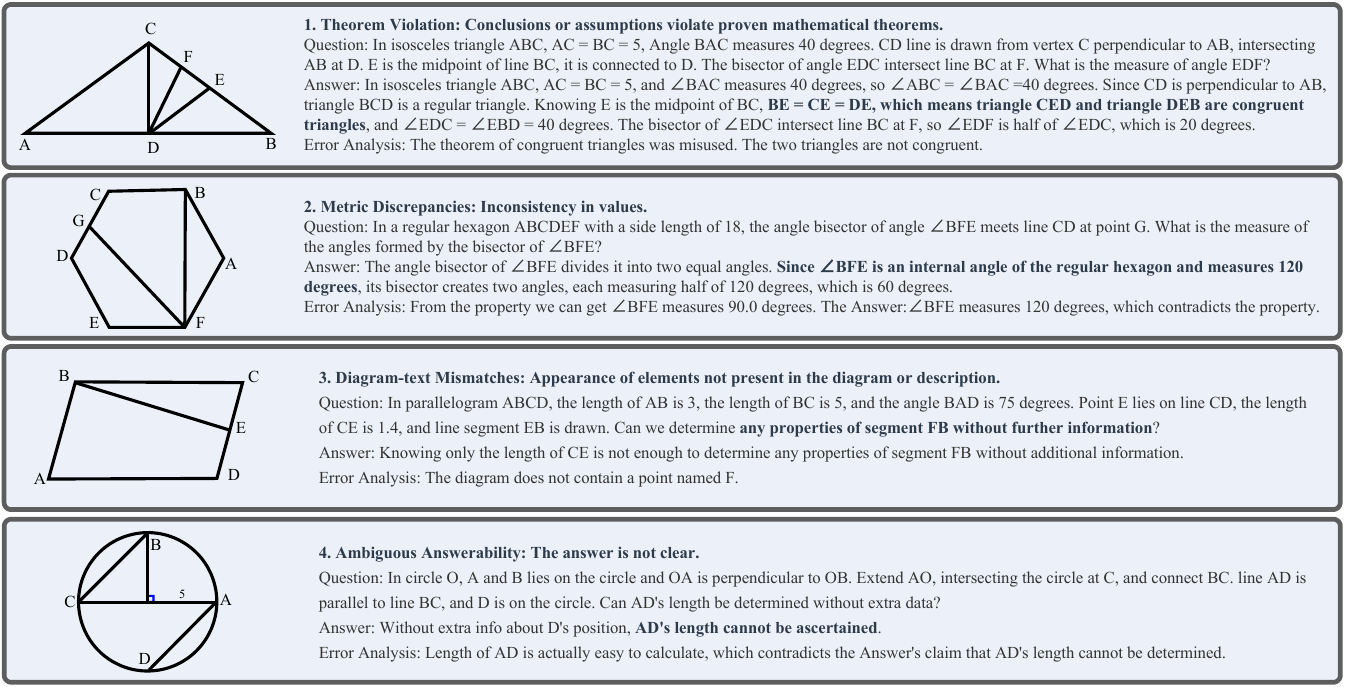}
    \caption{Examples of filtered errors.}
    \label{fig:error_filter_examples}
\end{figure*}

Fig.~\ref{fig:error_filter_examples} presents four representative types of errors identified by the Error A\&Q Filtering module. Theorem Violation refers to cases where conclusions or assumptions contradict established mathematical theorems. Metric Discrepancies involve inconsistencies between the given numerical values or angles and the geometric properties. Diagram-text Mismatches occur when elements described in the problem statement are either absent from the diagram or inconsistent with it. Ambiguous Answerability denotes problems in which the information provided is insufficient to derive a unique solution, or essential data is not explicitly stated in the question.

\section{Examples of TR-GeoSup dataset}
\label{app:geosup}
\begin{figure}[t!]
    \centering
    \includegraphics[width=0.9\linewidth]{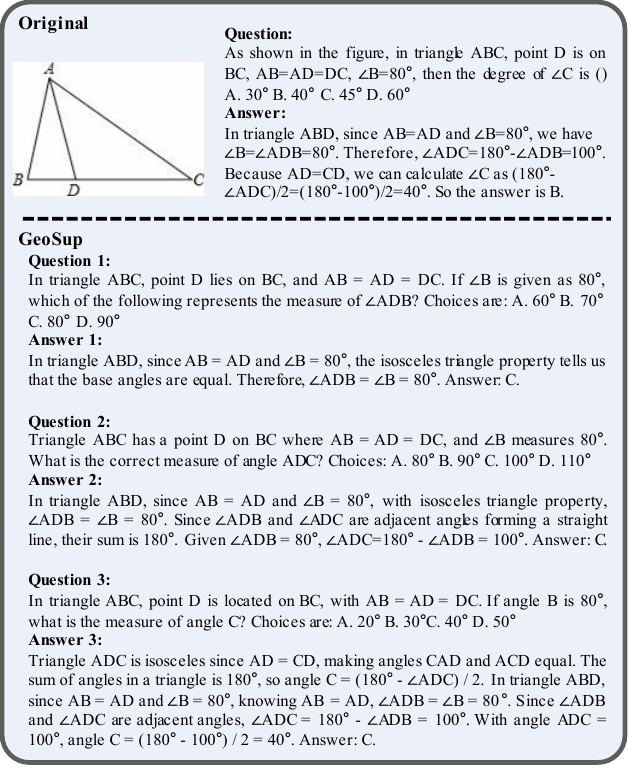}
    \caption{Examples of TR-GeoSup dataset.}
    \label{fig:GeoSup}
\end{figure}

Fig.~\ref{fig:GeoSup} illustrates an example from the TR-GeoSup dataset, showcasing the transformation of a multi-step reasoning problem from the original GeoQA dataset. In the original Q\&A pair, the reasoning process is condensed and lacks explicit intermediate steps, relying on implicit knowledge. TR-GeoSup decomposes the original reasoning process into three hierarchical sub-questions, each accompanied by a detailed and theorem-aware reasoning chain. This augmentation not only clarifies the implicit knowledge embedded in the original data but also provides a step-by-step guide for model training.

\section{Statistical Comparison With Related Datasets}
\label{app:compare}
\begin{table}[]
    \small
    \centering
    \caption{Statistic comparison between Geo170K, GeomVerse, and our data. En. Exis means Enhance Existing data. Fully syn. means Fully synthesized.`/'indicates the same number as GeoQA}
    \renewcommand\arraystretch{1.1}
    \setlength{\tabcolsep}{2.1mm}{
        \begin{tabular}{c|cccc}
        \hline
            Dataset name & Data type & Img. & Q\&A  & Theorem \\ \hline
            Geo170K & En. Exis. & 6.4K  & 110K & / \\
            GeomVerse & Fully syn. & 9.3K & 9.3K & 60 \\
            TR-GeoMM & Fully syn. & 15K & 45K & 110 \\
            TR-GeoSup & En. Exis. & 6.4K & 20K & / \\\hline
        \end{tabular}
    }
    \label{tab:data_compare}
\end{table}

Here we make a brief comparison between our proposed dataset with some related academic datasets.

\begin{itemize}[leftmargin=*,noitemsep]
    \item GeomVerse is a representative template-based method that generates geometrically oversimplified images by combining predefined polygons in fixed configurations. These images only contain polygon compositions and lack theorem-aware elements (e.g., midlines and angle bisectors). It has 9.3k synthetic images accompanied by Q\&A pairs, but their richness was limited by the absence of theorem-aware elements, covering only 60 theorems.
    \item Geo170K represents an augmented version of the existing GeoQA dataset. It primarily focuses on rephrasing Q\&A pairs, such as altering wording, swapping conditions and answers, or scaling numerical values while keeping the underlying theorems identical. This approach does not enhance the diversity of theorems covered in the dataset.
    \item TR-CoT enables theorem-driven multimodal reasoning by designing substrates and embedding theorem-aware elements based on theorem conditions, allowing generated images to support complex Q\&A construction. Unlike prior approaches, TR-CoT is not limited by existing data coverage and can expand a model's geometric knowledge. The framework supports both new data synthesis and augmentation of existing datasets, and covers 110 theorems through its structured generation process.
\end{itemize}

As shown in Tab.~\ref{tab:data_compare}, our data possess a notable diversity in theorem coverage, image distribution, and Q\&A quantity. Ablation studies in Section 4.2 further discuss the training effectiveness of our proposed data. 

\section{Detail of polygon distribution}
\label{app:geo_distribution}
We conducted robustness experiments for different polygon distributions, where the details of the polygon distributions are shown in Tab.~\ref{distributions_detail}. From top to bottom, the percentage of triangles and quads gradually decreases, and the percentage of pentagons and hexagons gradually increases. There is also a clear difference in the percentage of circles.

Similar quantitative results within 0.6\% in Tab.~\ref{distributions} show that the impact of polygon distributions is almost negligible, demonstrating the strong robustness of our method to different polygon distributions.
Therefore, the performance gain is mainly attributed to the diverse geometry representation and reasoning knowledge provided by our method. 

\begin{table}[h]
\small
\caption{Details of polygon distribution for distributional robust ablation studies.}
\centering
\renewcommand\arraystretch{1.1}
\begin{tabular}{m{.17\columnwidth}<{\centering}  | m{.09\columnwidth}<{\centering} m{.07\columnwidth}<{\centering} m{.07\columnwidth}<{\centering} m{.13\columnwidth}<{\centering} m{.13\columnwidth}<{\centering} }
\hline
\multirow{2}*{Method} & \multicolumn{5}{c}{Polygon Distribution} \\
\cline{2-6}
& triangle & quad & circle & pentagon & hexagon \\
\hline
Group I & 29\% & 46\% & 17\% & 5\% & 3\%  \\
Group II & 32\% & 40\% & 14\% & 8\% & 6\%  \\
Group III & 25\% & 33\% & 21\% & 12\% & 8\%  \\ \hline
\end{tabular}
\label{distributions_detail}
\end{table}

\begin{table}[h]
\small
\caption{Ablation study on the robustness to polygonal distributions.}
\centering
\renewcommand\arraystretch{1.1}
\setlength{\tabcolsep}{4mm}{
\begin{tabular}{c|cc}
\hline
Polygon Distribution & MathVista & GeoQA\\  \hline
Group I    & 64.4  & 54.0 \\
Group II   & 64.4  & 53.7 \\
Group III  & 63.9  & 53.4  \\ \hline
\end{tabular}
}
\label{distributions}
\end{table}

\section{The Case of Direct Generation and TR-Reasoner Generation}
\label{app:direct_reverse}
\begin{figure*}[t!]
    \centering
    \includegraphics[width=0.88\linewidth]{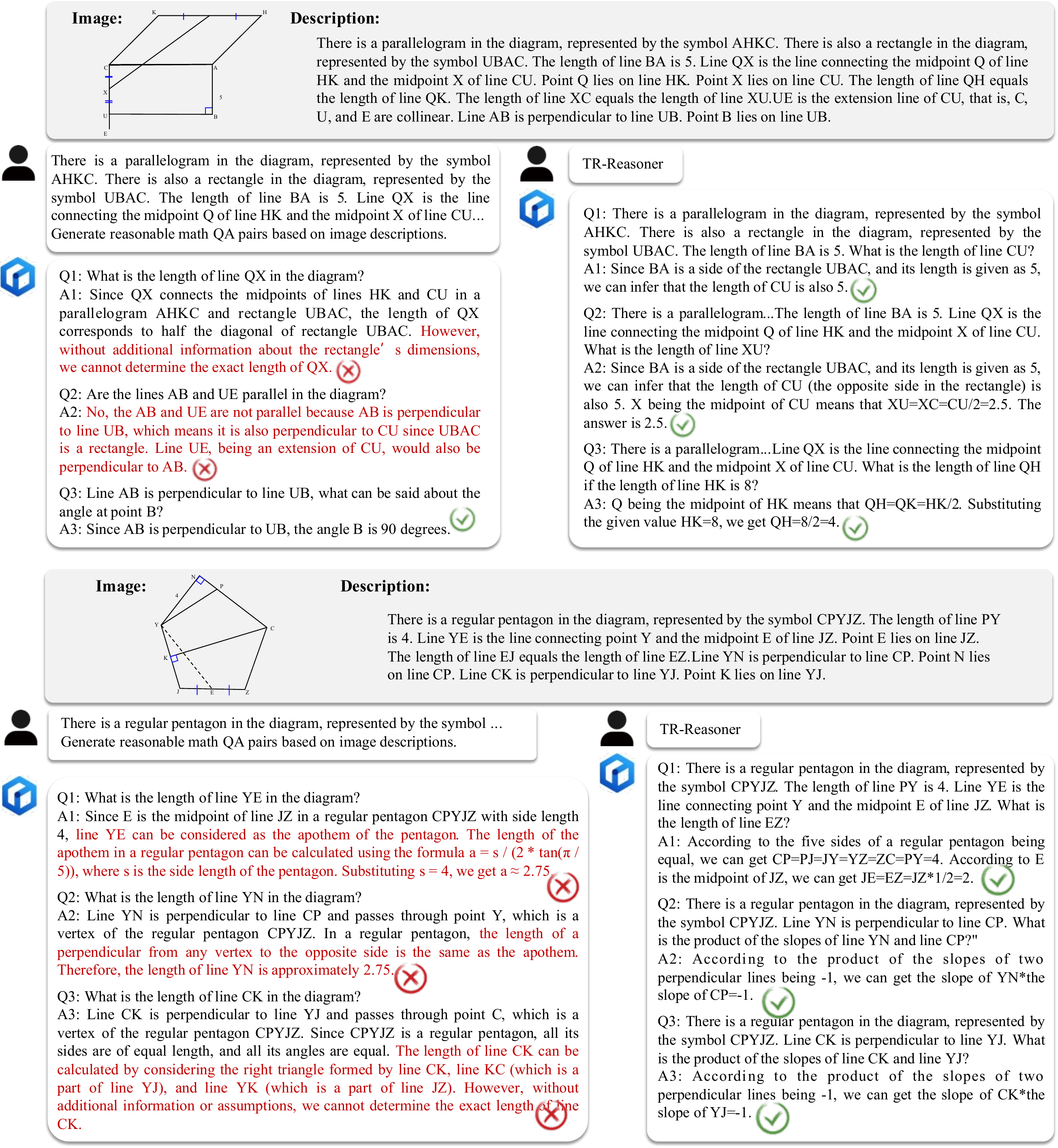}
    \caption{The Case of Direct Generation and TR-Reasoner Generation.}
    \label{fig:step_effect2}
\end{figure*}

The core idea of the TR-Reasoner is to improve the accuracy of Q\&A pairs by simplifying the reasoning based on descriptions and then generating corresponding questions from the answers in a reversed manner.
A straightforward approach is directly prompting ERNIE Bot 4.0 to generate Q\&A pairs from the input image description. However, as shown on the left of Fig.~\ref{fig:step_effect2}, this approach often fails to determine the correct answer.
In contrast, the Q\&A pairs produced by TR-Reasoner are correct for all three instances with our design.

\section{Details of the theorems}
\label{app:theorems}
The support of mathematical theorems is crucial for the accuracy of TR-Engine. In Tab.~\ref{geometry_theorems}, we present the geometric theorems and properties that we used. These define the rules for combining elements, establishing a logically coherent chain throughout the figure construction process. They serve as the foundation for extending reasoning scenarios and also assist in the computation and verification of question-answer pairs.

We collect and organize geometric theorems through three main approaches: 
\textbf{(1) Systematic Textbook Mining:} We analyzed standard textbooks and online educational resources to compile core geometric axioms and theorems from primary and secondary school mathematics curricula in Mainland China. 
\textbf{(2) Alignment with Public Academic Datasets:} We extracted theorems referenced in public academic datasets (e.g., PGPS9K, MAVIS, GeomVerse) to ensure consistency with commonly used training corpora. 
\textbf{(3) Expert Consultation:} We consulted primary and secondary school educators to identify important theorems and conclusions grounded in real-world teaching practices.

\section{Effectiveness of TR-CoT}
\label{app:Effectiveness}
As shown in Fig.~\ref{Effectiveness_result}, models jointly trained on Geo170K and TR-CoT-generated data (TR-GeoMM and TR-GeoSup) consistently outperform those trained solely on Geo170K (`Geo-'). 
InternVL2.5-8B receives a 1.5\% improvement on MathVista and GeoQA, and Qwen2.5-VL-7B improves by 1.0\% and 2.0\% on MathVista and GeoQA, respectively.
These results indicate that TR-CoT-generated data can supplement existing datasets and is widely effective in various LMMs.

\begin{table}[h]
\small
\centering
\caption{TR-CoT generated data effectiveness validation on different models. ‘Geo-’ indicates the model is fine-tuned only with geometric instruction data of Geo170K. Consistent and significant improvement without adding any additional parameters.}
\renewcommand\arraystretch{1.15}
\setlength{\tabcolsep}{2.7mm}{
\begin{tabular}{c|cc}
\hline
Model                       & MathVista         & GeoQA                          \\ \hline
Geo-InternVL-2.0-2B       & 51.9           &  62.5               \\ 
TR-CoT-InternVL-2.0-2B     & 56.3 \color{mydarkgreen}{(4.4$\uparrow$)}          &  63.4 \color{mydarkgreen}{(0.9$\uparrow$)}         \\ \hline
Geo-LLaVA-1.5-7B         &  27.9          &  47.6              \\
TR-CoT-LLaVA-7B           &  29.3 \color{mydarkgreen}{(1.4$\uparrow$)}        &  51.7 \color{mydarkgreen}{(4.1$\uparrow$)}          \\ \hline
Geo-Qwen2-VL-7B          &  59.9          &  69.1  \\ 
TR-CoT-Qwen2-VL-7B       & 67.6 \color{mydarkgreen}{(7.7$\uparrow$)}     & 70.4 \color{mydarkgreen}{(1.3$\uparrow$)}    \\ \hline
Geo-InternVL-2.0-8B      &  70.2          &  74.9 \\
TR-CoT-InternVL-2.0-8B       &  72.1 \color{mydarkgreen}{(1.9$\uparrow$)}    & 76.7 \color{mydarkgreen}{(1.8$\uparrow$)}   \\ \hline
Geo-InternVL-2.5-8B      &  76.4          &  75.2 \\
TR-CoT-InternVL-2.5-8B      &  77.9 \color{mydarkgreen}{(1.5$\uparrow$)}    &  76.7 \color{mydarkgreen}{(1.5$\uparrow$)} \\ \hline
Geo-Qwen2.5-VL-7B       &  73.5          &  77.2  \\
TR-CoT-Qwen2.5-VL-7B      &  74.5  \color{mydarkgreen}{(1.0$\uparrow$)}        &  79.2 \color{mydarkgreen}{(2.0$\uparrow$)} \\
\hline
\end{tabular}
  }
\label{Effectiveness_result}
\end{table}

\section{Details of CoT quality evaluation}
\label{app:cot_quality}
\begin{figure}[]
    \centering
    \includegraphics[width=1\linewidth]{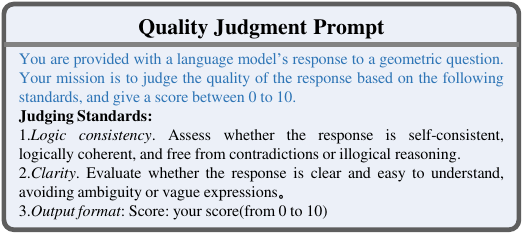}
    \caption{Comparison of model problem solving before and after training.}
    \label{fig:logical_prompt}
\end{figure}

We used ERNIE Bot 4.0 and DeepSeek R1 to evaluate model outputs. For each response, the evaluation model gives a score between 0 and 10 to judge the logical consistency, clarity, and lack of ambiguity. We use the average score of the two models as the final score. To ensure more accurate evaluation, we include specific judging standards. The prompts used are shown in Fig.~\ref{fig:logical_prompt}. The blue part represents the Task Description.

\begin{table*}[h]
\footnotesize 
\caption{Summary of Geometric Theorems and Properties}
\centering 
\begin{tabular}{p{0.3\columnwidth} p{0.7\columnwidth} p{0.7\columnwidth}} 
\hline
\textbf{Category} & \textbf{Properties} & \textbf{Criteria} \\
\hline
Parallel Lines & 
Corresponding angles equal; Alternate interior angles equal; Consecutive interior angles supplementary &
Equal corresponding angles; Supplementary consecutive angles; Equal alternate angles; Parallel to the same line \\
\hline

General Triangles & 
Interior angles sum to $180^\circ$ &
AA similarity; SSS/SAS/ASA/AAS/HL congruence \\
\hline

Isosceles Triangles & 
Equal base angles; Three-line coincidence (angle bisector, median, altitude) ;Base angles are $45^\circ$ in right-isosceles case &
Two equal angles ; Two equal sides \\
\hline

Equilateral Triangles & 
 All angles are $60^\circ$ ; Three - line coincidence &
 Three equal sides ; Three equal angles ; Isosceles triangle with a $60^\circ$ angle \\
\hline

Right Triangles & 
 Acute angles are complementary ; Side opposite $30^\circ$ angle is half of the hypotenuse ; Median on the hypotenuse is half of the hypotenuse ; Pythagorean theorem: $a^2 + b^2 = c^2$ &
 Contains a right angle ; HL congruence for right - triangles \\
\hline

Angle Bisector & 
 Points on the perpendicular bisector are equidistant from the endpoints &
 A ray that divides an angle into two equal parts \\
\hline

Triangle Midline & 
 Parallel to the third side and half of its length &
 Connects the mid-points of two sides \\
\hline

Parallelogram & 
 Opposite sides are equal ; Diagonals bisect each other ; Area $= base\times height$ &
 Both pairs of opposite sides are parallel; Diagonals bisect each other; Opposite sides are equal \\
\hline

Rectangle & 
 All angles are $90^\circ$ ; Diagonals are equal &
 A parallelogram with a right angle; A quadrilateral with three right angles \\
\hline

Rhombus & 
 All sides are equal; Diagonals are perpendicular to each other &
 A parallelogram with adjacent sides equal; A quadrilateral with four equal sides \\
\hline

Square & 
 All sides and angles are equal; Diagonals are equal and perpendicular &
 Prove it is both a rectangle and a rhombus \\
\hline

Isosceles Trapezoid & 
 Legs are equal; Base angles on the same base are equal &
 Two equal legs; Equal base angles on the same base \\
\hline

Trigonometric Functions & 
 $\sin30^{\circ}=\frac{1}{2}$ ; $\sin45^{\circ}=\frac{\sqrt{2}}{2}$ ; $\sin60^{\circ}=\frac{\sqrt{3}}{2}$ ; $\sin90^{\circ}=1$ ; $\cos30^{\circ}=\frac{\sqrt{3}}{2}$ ; $\cos45^{\circ}=\frac{\sqrt{2}}{2}$ ; $\cos60^{\circ}=\frac{1}{2}$ ; $\cos90^{\circ}=0$ ; $\tan30^{\circ}=\frac{\sqrt{3}}{3}$ ; $\tan45^{\circ}=1$ ; $\tan60^{\circ}=\sqrt{3}$ &
/ \\
\hline

Circle & 
 The perpendicular bisector of a chord is perpendicular to the chord; The perpendicular bisector of a chord passes through the center &
/ \\
\hline

Central Angle & 
 Equal central angles subtend equal chords and arcs &
/ \\
\hline

Inscribed Angle & 
 An inscribed angle is half of the central angle subtended by the same arc; An angle subtended by a diameter is a right angle &
/ \\
\hline

Cyclic Quadrilateral & 
 Opposite angles are supplementary &
/ \\
\hline

Tangent & 
 A tangent is perpendicular to the radius at the point of contact; Tangents from an external point to a circle are equal in length &
 A line perpendicular to the radius at the endpoint on the circle is a tangent \\
\hline

Regular Polygon & 
 For an equilateral triangle inscribed in a circle of radius $R$, side length $a = R\sqrt{3}$ ; For a square inscribed in a circle of radius $R$, side length $a = R\sqrt{2}$ &
/ \\
\hline
\end{tabular}%
\label{geometry_theorems}
\end{table*}

\end{document}